\documentclass[letterpaper, 10 pt, conference]{ieeeconf}  

\IEEEoverridecommandlockouts                              

\usepackage{epsfig} 
\usepackage{threeparttable}
\usepackage{booktabs}
\usepackage{graphics} 
\usepackage{amsmath} 
\usepackage{amssymb}  
\usepackage{xcolor}
\usepackage{array}
\usepackage{multirow}
\usepackage{cite}
\usepackage{subcaption}
\usepackage{graphicx}
\usepackage{adjustbox} 
\usepackage{svg}
\usepackage{cuted}

\title{\LARGE \bf
FTSplat: Feed-forward Triangle Splatting Network
}

\author{Jinlin Xiong, Can Li, Jiawei Shen, Zhigang Qi, Lei Sun$^{*}$, Dongyang Zhao}

\begin{document}
\maketitle
\thispagestyle{empty}
\pagestyle{empty}

\vspace{-6mm}
\begin{strip}
\centering
\includegraphics[width=0.95\textwidth]{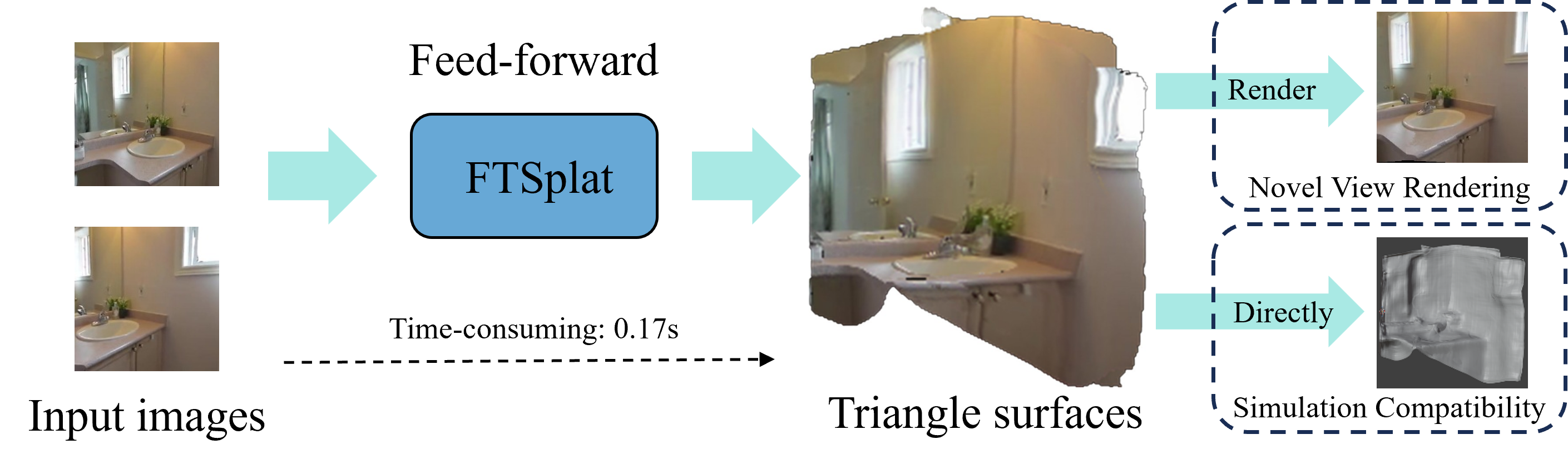}
\captionof{figure}{Overview of the proposed FTSplat.
Given multi-view input images, our feed-forward FTSplat directly and efficiently predicts a triangular surface representation of the scene. 
The reconstructed mesh supports photo-realistic novel view rendering and can be readily imported into simulation software such as Blender for downstream applications.
Compared to existing optimization-based triangular surface methods that typically require several minutes for reconstruction, our approach enables scene modeling within sub-second. 
}
\label{fig:overview}
\end{strip}

\footnotetext[1]{This work was supported in part by the National Natural Science Foundation of China under grant 62173192 and in part by the Shenzhen Science and Technology Program under grant JCYJ20220530162202005.}

\footnotetext[2]{Jinlin Xiong, Can Li, Jiawei Shen and Lei Sun are with Tianjin Key Laboratory of Intelligent Robotics, Nankai University, Tianjin 300350, China (tjKLIR).}

\footnotetext[3]{Dongyang Zhao and Zhigang Qi are with Beijing Institute of Computer Application, Beijing 100081, China.}

\begin{abstract}


High-fidelity three-dimensional (3D) reconstruction is essential for robotics and simulation. 
While Neural Radiance Fields (NeRF) and 3D Gaussian Splatting (3DGS) achieve impressive rendering quality, their reliance on time-consuming per-scene optimization limits real-time deployment. Emerging feed-forward Gaussian splatting methods improve efficiency but often lack explicit, manifold geometry required for direct simulation.
To address these limitations, we propose a feed-forward framework for triangle primitive generation that directly predicts continuous triangle surfaces from calibrated multi-view images. Our method produces simulation-ready models in a single forward pass, obviating the need for per-scene optimization or post-processing. 
We introduce a pixel-aligned triangle generation module and incorporate relative 3D point cloud supervision to enhance geometric learning stability and consistency.
Experiments demonstrate that our method achieves efficient reconstruction while maintaining seamless compatibility with standard graphics and robotic simulators.

\end{abstract}

\section{INTRODUCTION}
High-fidelity 3D reconstruction is a fundamental capability for robotic perception, simulation, and digital twin applications.
As robotic systems increasingly require accurate environmental understanding and physically consistent scene representations, the efficient and accurate recovery of 3D scene models from images has become a fundamental problem in robotics and computer vision.
Recently, Neural radiance fields (NeRF)~\cite{nerf} methods have demonstrated remarkable progress by representing scenes with implicit continuous functions, significantly improving novel view synthesis and reconstruction quality. Building upon this line of work, 3D Gaussian Splatting (3DGS)~\cite{3dgs} further introduces explicit Gaussian primitives to enable high-quality and real-time rendering, substantially advancing the practical deployment of 3D reconstruction systems.

However, most of NeRF-based and 3DGS-based methods rely on per-scene optimization or iterative procedures, which significantly limits their deployment efficiency in large-scale scene reconstruction and online robotic applications.
To address this limitation, recent studies have explored feed-forward Gaussian Splatting approaches~\cite{pixelsplat, mvsplat, depthsplat} that directly predict Gaussian primitives from multi-view images in a single forward pass, achieving substantially improved inference efficiency while maintaining competitive reconstruction quality.


Despite the advantages of 3DGS in rendering performance, Gaussian primitives lack explicit geometric structure, making them difficult to directly integrate with mainstream physics-based simulators and robotic simulation platforms. 
To bridge this gap, recent mesh-based representation methods~\cite{meshsplatting, radiance_mesh, trianglesplatting} replace Gaussian primitives with triangular facets for scene representation, offering more standardized and simulation-compatible geometric structures while preserving efficient rendering.
Such explicit mesh representations can be directly imported into widely used simulation and graphics software, such as Blender, without requiring additional surface reconstruction or post-processing.
However, their reliance on iterative optimization introduces significant computational cost, limiting scalability and reducing practicality in time-sensitive robotic applications.

To bridge the gap between feed-forward efficiency and explicit geometric representation, we propose a feed-forward triangular primitive generation framework that directly predicts continuous triangular surface primitives from multi-view images.
By combining the efficiency of feed-forward reconstruction with the structural advantages of mesh-based representations, our method generates simulation-compatible surface structures in a single forward pass. 
This enables efficient and high-fidelity 3D reconstruction while producing continuous triangular surfaces that can be directly imported into existing graphics and robotics simulation environments, such as Blender, without additional mesh reconstruction or post-processing, thereby facilitating seamless integration with simulation and digital twin pipelines.

Specifically, we present a feed-forward framework that directly generates a triangular surface representation of a 3D scene from multiple images. Given a set of calibrated multi-view images, our method extracts image features using a pretrained vision transformer and a multi-view transformer encoder. 
Based on the feature matching strategy~\cite{mvsplat},
a sparse feature point cloud is constructed and subsequently decoded into a set of 3D vertices augmented with opacity and color represented by spherical harmonics.
The resulting point-based representation is further processed by a pixel-aligned triangular surface generation module, which predicts the face connectivity and constructs an explicit triangular surface representation. The generated triangular primitives are rendered via a differentiable triangle rasterizer~\cite{trianglesplatting}, and the rendered images are supervised using photometric losses against the ground-truth views. In addition, since triangular surface reconstruction critically depends on accurate 3D geometry, we introduce an auxiliary 3D point cloud supervision during early training stages with a higher weight to encourage rapid geometric convergence. As training progresses, the weight of the 3D supervision is gradually reduced, allowing the optimization to focus on high-quality appearance and texture reconstruction.

In summary, the main contributions of this paper include:

\begin{itemize}
\item We propose the first feed-forward framework that directly predicts continuous triangular surface representations of 3D scenes from multi-view images, which are natively compatible with existing graphics and robotics simulation platforms. This enables efficient generation of continuous surface models without per-scene optimization or additional post-processing.

\item We design a pixel-aligned triangular surface generation module that converts feature point clouds predicted by the feed-forward network into explicit triangular surface primitives suitable for efficient rasterization.

\item We introduce a relative 3D point cloud supervision and adopt a geometry-to-appearance training strategy, which emphasizes geometric consistency in the early training stage and gradually shifts the optimization focus toward high-quality texture reconstruction, enabling stable and accurate convergence of the proposed framework.
\end{itemize}

\section{RELATED WORK}

\begin{figure*}[thpb]
    \centering
    \vspace{2mm}
    \includegraphics[scale=0.6]{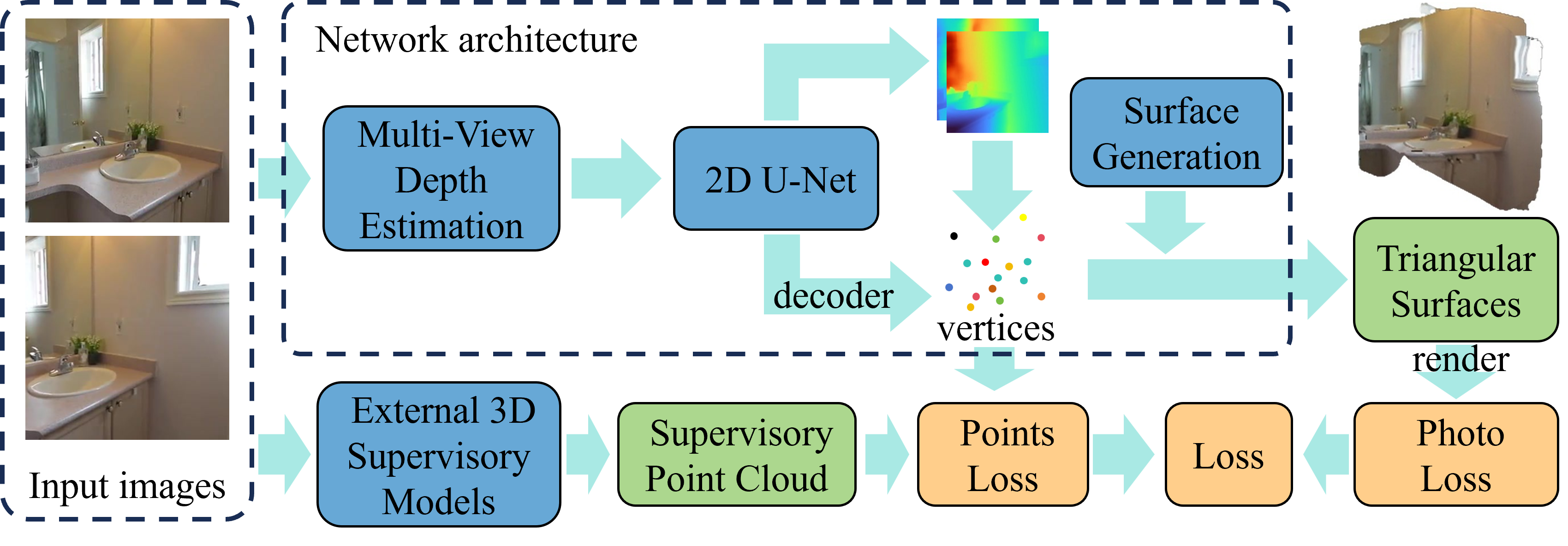}
    \caption{Overview of the proposed feed-forward triangular surface reconstruction network. 
    Multi-view images are processed by a Multi-View Depth Estimation module to obtain fused features enriched with depth information.
    The fused features are used to predict depth maps and back-project an initial 3D point cloud, while a 2D U-Net with a triangle head decodes additional vertex attributes (opacity and spherical harmonics color). A surface generation module infers face connectivity to produce the final triangular surface. Differentiable rasterization enables photometric supervision, and external 3D point cloud supervision provides explicit geometric constraints during training.}
    \label{fig1}
\end{figure*}

\subsection{Optimization-based 3D Reconstruction Methods}
Optimization-based 3D reconstruction methods have long been the dominant paradigm for high-fidelity scene reconstruction. These approaches typically rely on explicit or implicit per-scene parameter optimization, aiming to minimize multi-view consistency or reprojection errors to recover detailed geometric and appearance representations. Neural Radiance Field (NeRF) methods represent scenes as continuous implicit functions and optimize them via volumetric rendering, achieving remarkable progress in novel view synthesis and high-quality reconstruction.
Subsequent works have focused on improving the training efficiency and representational capacity of NeRF, including approaches such as Instant-NGP~\cite{instantngp}, which significantly accelerates optimization through multi-resolution hash encoding, as well as various variants that explore alternative sampling strategies, network architectures, and regularization schemes~\cite{plenoxels, mip360, dsnerf}. While these methods strike different trade-offs between reconstruction quality and computational efficiency, they generally rely on per-scene optimization, which limits their scalability to large-scale environments and real-time applications.

To address the aforementioned limitations, 3D Gaussian Splatting introduces an explicit scene representation based on Gaussian primitives. By directly optimizing the spatial parameters and appearance attributes of Gaussians, 3DGS enables high-quality reconstruction with efficient real-time rendering. Building upon this framework, subsequent studies have proposed a variety of extensions and improvements in terms of geometric modeling, appearance representation, and training stability, such as regularizing Gaussian distributions, introducing hierarchical representations, and extending the formulation to dynamic or editable scenes~\cite{HRGS, 3dgs2, ralgs}. While these methods further enhance reconstruction quality and broaden the applicability of Gaussian-based representations, their core pipelines still rely on scene-specific optimization procedures.
More recently, 2D Gaussian Splatting~\cite{2dgs} reformulates the representation by anchoring Gaussian primitives in image-aligned or surface-aligned two-dimensional domains, which improves geometric consistency and reduces ambiguities caused by unconstrained 3D Gaussians. Nevertheless, 2DGS remains a scene-specific optimization-based approach, requiring per-scene training to obtain high-quality reconstructions.

Recent studies have further explored Mesh Splatting methods that adopt triangular facets as explicit scene representations. By replacing Gaussian primitives with triangular elements or explicit surface primitives, these approaches retain the rendering efficiency of splatting-based pipelines while introducing more standardized geometric representations. Compared to Gaussian-based formulations, triangular meshes offer superior compatibility with mainstream graphics and physics simulation engines, facilitating the direct use of high-quality reconstructed models in downstream tasks such as simulation, collision detection, and physical dynamics analysis. However, existing Mesh Splatting methods still rely on per-scene optimization strategies, resulting in considerable computational cost and limited deployment efficiency, which remain insufficient for online robotic applications.

\subsection{Feed-forward 3D Reconstruction Methods}
In contrast to optimization-based approaches, feed-forward 3D reconstruction methods aim to directly predict scene representations from multi-view images in a single forward pass, thereby significantly improving inference efficiency and system scalability.

PixelSplat~\cite{pixelsplat} is among the earliest works to explore feed-forward Gaussian Splatting, where a neural network directly predicts Gaussian primitives from multi-view images, enabling fast 3D reconstruction without per-scene optimization. Building upon PixelSplat, Mvsplat~\cite{mvsplat} introduces multi-view feature matching and explicit geometric constraints, significantly improving geometric consistency and reconstruction quality of the feed-forward predicted Gaussians. Furthermore, Depthsplat~\cite{depthsplat} incorporates monocular depth features into the feed-forward Gaussian Splatting framework, leveraging depth priors to enhance geometric recoverability and achieve more stable reconstruction results in complex scenes.

Despite the significant improvements in inference efficiency achieved by the aforementioned approaches, they continue to rely on Gaussian primitives as scene representations, which lack explicit geometric surface structures and thus limit their direct applicability to simulation and robotic tasks. In contrast, our method further explores a feed-forward triangular surface generation strategy that preserves the efficiency of feed-forward inference while introducing explicit triangular surface representations, enabling high-fidelity 3D reconstruction that is more compatible with physics-based simulation platforms.

\section{ALGORITHM}

This section provides a detailed description of the proposed approach. 
In Section~\ref{sec:31}, we review the necessary background, including the definition of triangular surface representations and the rendering process of triangular primitives. 
In Section~\ref{sec:32}, we present the technical details of our method, covering the overall network architecture, loss function design, and training strategy.


\subsection{Preliminaries}
\label{sec:31}
To ensure better compatibility with existing graphics and simulation platforms, the triangular surface representation generated by our feed-forward network follows the formulation of MeshSplatting~\cite{meshsplatting}, namely a continuous triangular surface model.
Specifically, the model consists of a set of vertices and their associated attributes, including vertex positions $\mathbf{v}$, opacity values $o$, RGB colors $\mathbf{c}$ represented using spherical harmonics (SH) coefficients, and a smoothing parameter $\sigma$. The corresponding dimensions are $\mathbf{v} \in \mathbb{R}^{N \times 3}$, $o \in \mathbb{R}^{N}$, and $\mathbf{c} \in \mathbb{R}^{N \times 3 \times d_h}$. In our implementation, the smoothing parameter $\sigma$ is set to zero by default.
In addition to vertex attributes, the triangular surface model includes a face connectivity tensor $\mathbf{f} \in \mathbb{R}^{T \times 3}$, where each row stores the indices of the three vertices forming a triangular face. This connectivity explicitly defines a surface composed of $T$ triangular primitives.

Based on the triangular surface representation described above, we employ a differentiable triangle rasterization process to render the reconstructed geometry. Specifically, the triangular vertices $\mathbf{v}$ are first projected onto the 2D image plane using a standard pinhole camera model, yielding their corresponding positions in pixel space. Since the smoothing parameter $\sigma$ is fixed to zero in our formulation, the reconstructed triangular surfaces exhibit explicit and well-defined boundaries, without introducing additional boundary smoothing or soft blending effects.
The rendering strategy of the triangular primitives follows the same formulation as that adopted in 3D Gaussian Splatting, which can be expressed as follows:
\begin{equation}
\label{eq1}
C(\mathbf{p})=\sum_{i=1}^{n} c_i o_i \prod_{j=1}^{i-1}\left(1- o_j\right)
\end{equation}

\subsection{Feed-forward Triangular Surface Reconstruction}
\label{sec:32}
Figure \ref{fig1} illustrates the overall architecture of our feed-forward network. Given $n$ multi-view input images, we first extract image features and predict depth maps using a multi-view depth estimation module. The features are then processed by a 2D U-Net followed by a triangle head to decode the vertex representations. Finally, a Surface Generation module is applied to infer face connectivity, producing the final triangular surface model.

\textbf{Feature Extraction.}
We first extract image features using a lightweight ResNet~\cite{resnet} with shared weights across views. The resulting features are then processed by a six-layer multi-view Swin Transformer~\cite{swintrans, gmflow, uni_flow} to exchange information among different views, producing multi-view features $\{ F^i_{mv} \}^n_{i=1}, F^i_{mv}\in\mathbb{R}^{\frac{H}{s} \times \frac{W}{s} \times C_1}$. 
In addition, we employ a pretrained Depth Anything V2~\cite{da2} model to extract monocular depth-aware features $\{ F^i_{mono} \}^n_{i=1}, F^i_{mv} \in \mathbb{R}^{\frac{H}{s} \times \frac{W}{s} \times C_2}$.
The two types of features are subsequently fused to form the final image feature representation.

\textbf{Multi-View Depth Estimation.}
We estimate depth maps for the input images using a cost volume based approach. Specifically, we uniformly sample $D$ depth hypotheses within a predefined near and far depth range. For the feature map $F_j$ of view $j$, we leverage the known camera intrinsics and extrinsics to project the features onto the image plane of view $i$ at each depth hypothesis, resulting in $D$ warped features ${F_{j \rightarrow i}}$. The warped features ${F_{j \rightarrow i}}$ are then compared with the reference view features $F_i$ using a dot-product operation to compute feature correlations, which are aggregated to construct the cost volume ${C_i \in \mathbb{R}^{\frac{H}{s} \times \frac{W}{s} \times D}}$. Finally, the cost volume is used as weights to compute a weighted sum over the $D$ depth hypotheses, yielding the final depth estimate for each view.

\textbf{Triangular Surface Generation.}
The constructed cost volume, multi-view features, monocular features, and predicted depth maps are jointly fed into a 2D U-Net~\cite{unet} to produce per-pixel feature representations. Meanwhile, using the estimated depth maps and known camera intrinsics and extrinsics, image pixels are back-projected into 3D space to obtain an initial point cloud.
We then employ a lightweight multi-layer perceptron (MLP), referred to as the triangle head, to decode per-point attributes, including opacity $o$ and color represented by spherical harmonics (SH) coefficients. This process yields the vertices of the triangular surface representation along with their associated attributes.

To handle depth discontinuities, we initially explored depth-gradient-based edge pruning and 3D KNN connectivity in Euclidean space. However, empirical results show that these strategies tend to introduce uncontrollable holes or break surface continuity. In contrast, direct pixel-level connectivity achieves the highest computational efficiency and provides more stable global topology in practice.

To construct the surface connectivity, we further adopt a simple yet effective pixel-level face generation strategy. Specifically, the number of generated points is ${V = n \times \frac{H}{s} \times \frac{W}{s}}$, where $n$ denotes the number of input views, $H$ and $W$ are the input image resolution, and $s$ is the downsampling factor. Each 3D point is mapped back to its corresponding image pixel coordinate ${(u, v)}$. For each pixel, we generate two adjacent triangular faces by connecting the vertices corresponding to neighboring pixels ${(u+1, v), (u, v+1)}$ and ${(u-1, v), (u, v-1)}$, respectively. This strategy ensures full coverage of the visible surface while minimizing the number of generated triangular faces, resulting in a compact and efficient triangular surface representation.

\begin{figure*}[t]
\centering
\vspace{2mm}
\setlength{\tabcolsep}{2pt} 
\renewcommand{\arraystretch}{1.0}

\begin{tabular}{c c c c c}

\textbf{Input images} &
\textbf{trianglesplatting} &
\textbf{meshsplatting} &
\textbf{FTSplat (Ours)} &
\textbf{GT} \\


\includegraphics[width=0.1\linewidth]{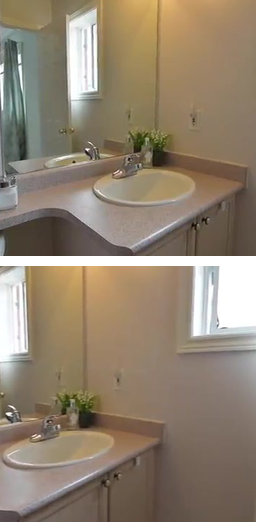} &
\includegraphics[width=0.2\linewidth]{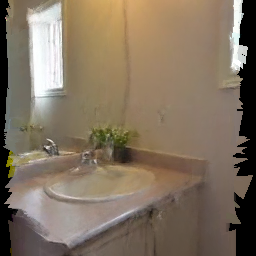} &
\includegraphics[width=0.2\linewidth]{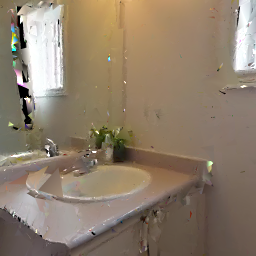} &
\includegraphics[width=0.2\linewidth]{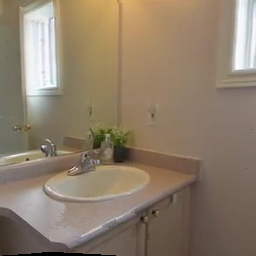} &
\includegraphics[width=0.2\linewidth]{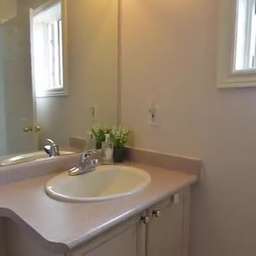} \\

\includegraphics[width=0.1\linewidth]{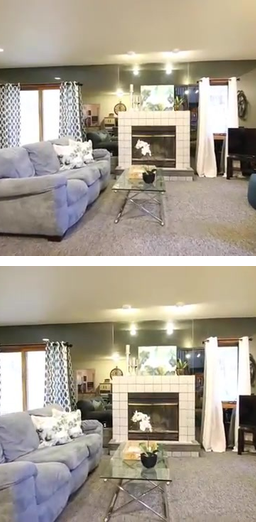} &
\includegraphics[width=0.2\linewidth]{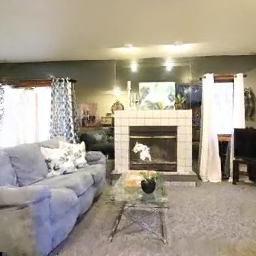} &
\includegraphics[width=0.2\linewidth]{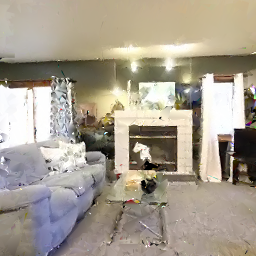} &
\includegraphics[width=0.2\linewidth]{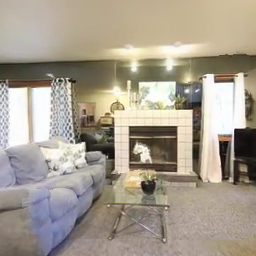} &
\includegraphics[width=0.2\linewidth]{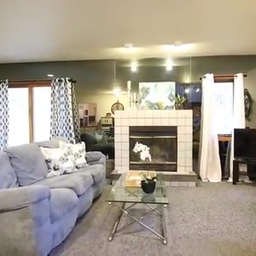} \\

\includegraphics[width=0.1\linewidth]{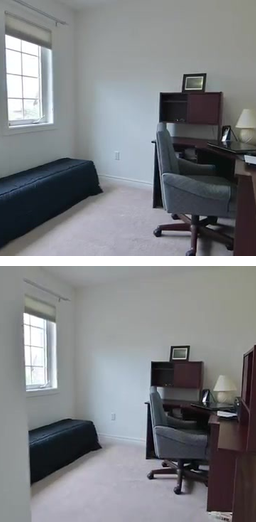} &
\includegraphics[width=0.2\linewidth]{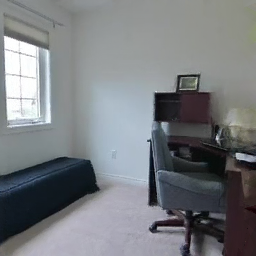} &
\includegraphics[width=0.2\linewidth]{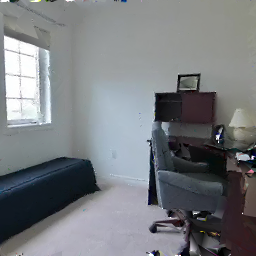} &
\includegraphics[width=0.2\linewidth]{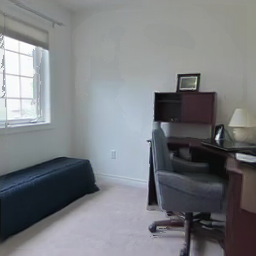} &
\includegraphics[width=0.2\linewidth]{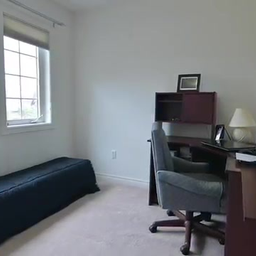} \\


\end{tabular}

\caption{
Qualitative comparison of reconstruction quality between our method and optimization-based triangle rasterization methods.
}
\label{fig_triangle}
\end{figure*}

\begin{figure*}[t]
\centering
\vspace{2mm}

\setlength{\tabcolsep}{3pt}
\renewcommand{\arraystretch}{1.0}

\newlength{\imgH}
\setlength{\imgH}{3.2cm}

\begin{tabular}{ccccc}

\multicolumn{2}{c}{\textbf{Input images}} &
\textbf{Mvsplat} &
\textbf{Depthsplat} &
\textbf{FTSplat (Ours)} \\

\includegraphics[height=\imgH]{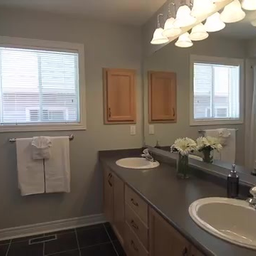} &
\includegraphics[height=\imgH]{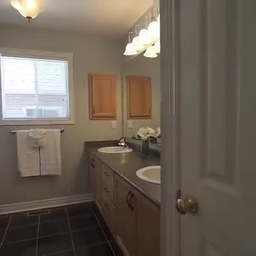} &
\includegraphics[height=\imgH]{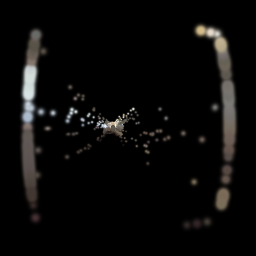} &
\includegraphics[height=\imgH]{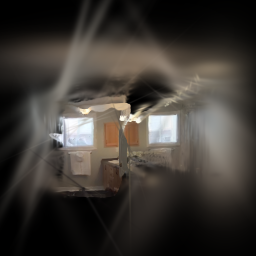} &
\includegraphics[height=\imgH]{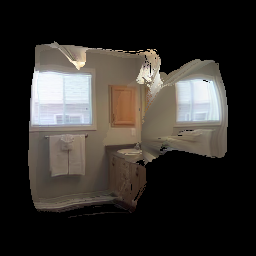} \\


\includegraphics[height=\imgH]{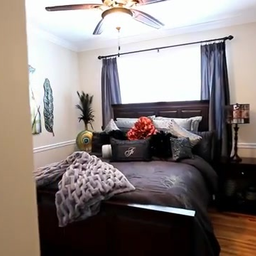} &
\includegraphics[height=\imgH]{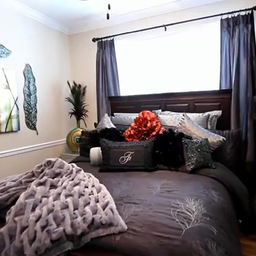} &
\includegraphics[height=\imgH]{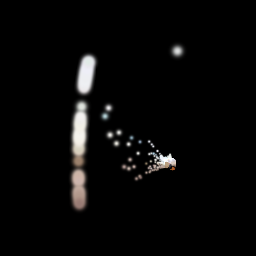} &
\includegraphics[height=\imgH]{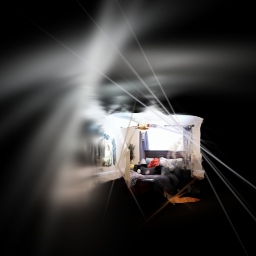} &
\includegraphics[height=\imgH]{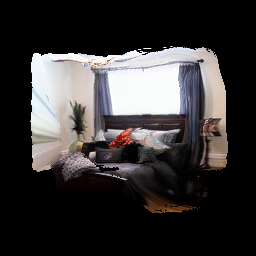} \\

\includegraphics[height=\imgH]{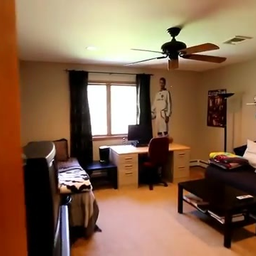} &
\includegraphics[height=\imgH]{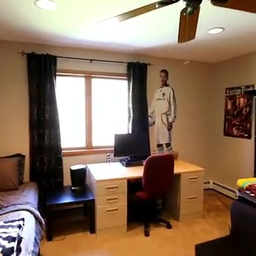} &
\includegraphics[height=\imgH]{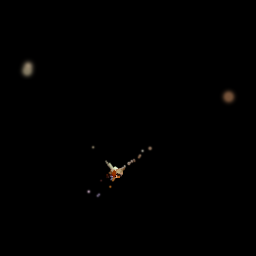} &
\includegraphics[height=\imgH]{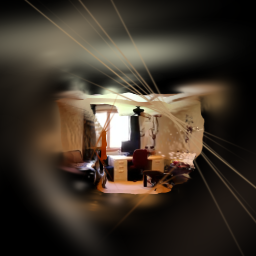} &
\includegraphics[height=\imgH]{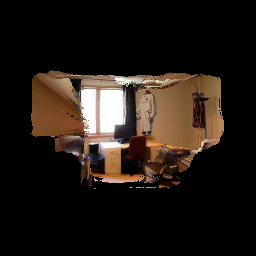} \\

\end{tabular}

\caption{
Qualitative comparison of 3D spatial consistency between our method and feed-forward Gaussian Splatting methods.
}
\label{fig_gs}
\end{figure*}

\textbf{Loss Function Design.}
The overall loss function of our method consists of two components: an image-rendering-related photometric loss and a geometry-related loss defined on the reconstructed 3D point cloud. The formula is as follows:
\begin{equation}
\label{eq2}
\mathcal{L} = \mathcal{L}_{photo} + \lambda_{points} \mathcal{L}_{points}
\end{equation}

The photometric loss ${\mathcal{L}_{photo}}$ is defined as:
\begin{equation}
\label{eq3}
\mathcal{L}_{photo}=\mathcal{L}_{L1} + \lambda_{lpips} \mathcal{L}_{LPIPS} + \lambda_{ds} \mathcal{L}_{ds}
\end{equation}

where ${\mathcal{L}_{L1}}$ denotes the $L1$ difference between the rendered image and the corresponding ground truth image, and ${\mathcal{L}_{LPIPS}}$ represents the perceptual similarity loss between the rendered image and the ground-truth image measured by LPIPS~\cite{lpips}.
In addition, we introduce a depth smoothness loss ${\mathcal{L}_{ds}}$~\cite{depthsmooth}, formulated as:
\begin{equation}
\label{eq4}
\mathcal{L}_{ds} = |\partial_x D_{render}| e^{-|\partial_x I_{gt}|}
+ |\partial_y D_{render}| e^{-|\partial_y I_{gt}|}
\end{equation}

This term serves as a regularization on the estimated depth maps, encouraging the depth gradients to be consistent with the image gradients, thereby promoting smooth depth variations while preserving sharp depth discontinuities at image edges.

Given that triangular surface representations require stronger 3D geometric consistency than Gaussian-based methods, we introduce an additional geometry-aware loss defined on the reconstructed 3D point cloud.
Specifically, we use multi-view 3D point clouds predicted by multiple foundation depth models, including Depth Anything V3~\cite{da3} and VGGT~\cite{vggt}, as supervisory signals to guide training. The geometric loss is defined as:
\begin{equation}
\label{eq5}
\mathcal{L}_{\text{points}}
=
\frac{1}{B}
\sum_{b=1}^{B}
\frac{1}{N}
\sum_{i=1}^{N}
\left\|
\mathcal{N}\!\left(\mathbf{V}_{b,i}\right)
-
\mathcal{N}\!\left({\mathbf{P}}_{b,i}\right)
\right\|_2^2 ,
\end{equation}

where ${\mathbf{V}}$ denotes the triangular vertices predicted by the proposed feed-forward network, and ${\mathbf{P}}$ represents the 3D point cloud predicted by Depth Anything V3. $\mathcal{N}(\cdot)$ denotes a robust normalization operator that removes global translation and scale ambiguity, which is defined as follows:
\begin{equation}
\label{eq6}
\mathcal{N}(\mathbf{X}_{b,i})
=
\frac{
\mathbf{X}_{b,i}
-
\operatorname{median}(\mathbf{X}_b)
}{
Q_{\alpha}
\!\left(
\left\|
\mathbf{X}_{b,i}
-
\operatorname{median}(\mathbf{X}_b)
\right\|_2
\right)
}
\end{equation}

where $Q_{\alpha}(\cdot)$ denotes the $\alpha$-quantile operator, which suppresses the influence of outliers by using the quantile of the distance distribution as a robust scale estimator.

Specifically, since the supervisory point clouds provide only relative geometric structure without absolute scale information, we impose the geometric constraint in a relative coordinate space on the vertices predicted by our feed-forward network. This encourages the reconstructed triangular surface representations to be more compact and effectively alleviates the floating primitives near object boundaries that are commonly observed in feed-forward Gaussian Splatting methods, thereby improving geometric consistency in 3D space.

Furthermore, to accelerate training convergence, we assign a larger weight ${\lambda_{points}}$ to the geometry loss during the early training stage, allowing the network to focus on learning stable 3D geometric structures. As training progresses, the weight ${\lambda_{points}}$ is gradually reduced, enabling the image-rendering-related photometric loss to dominate the optimization and guiding the network to emphasize high-quality texture and appearance reconstruction on triangular surfaces.

\section{EXPERIMENT}

We conduct experiments on the RealEstate10K~\cite{re10k} dataset at a resolution of $256 \times 256$. Our FTSplat model is trained on the RealEstate10K dataset for 400k iterations with a batch size of 2. Using identical reconstruction and evaluation viewpoints, we first compare our method with existing optimization-based triangle rasterization approaches, including Triangle Splatting~\cite{trianglesplatting} and MeshSplatting~\cite{meshsplatting}. In addition, we include 3D Gaussian Splatting (3DGS)~\cite{3dgs} and 2D Gaussian Splatting (2DGS)~\cite{2dgs} as representative Gaussian-based baselines. For all methods, each scene is reconstructed from two input views and evaluated on three target views.
Reconstruction quality is quantitatively evaluated using PSNR, SSIM~\cite{ssim}, and LPIPS~\cite{lpips}.
Furthermore, we compare FTSplat with several representative feed-forward Gaussian-based methods~\cite{mvsplat, depthsplat} to evaluate reconstruction quality. Finally, we perform ablation studies to validate the effectiveness of the proposed geometry-related loss.

\subsection{Reconstruction Quality Comparison with Optimization-based Methods}
In this section, we compare our feed-forward triangle surfaces generation approach with existing optimization-based triangle rasterization methods. In addition, conventional 3D Gaussian Splatting and 2D Gaussian Splatting are included as reference baselines.
Since optimization-based methods require scene-specific iterative refinement and incur substantial computational cost, we conduct experiments on a subset of several dozen scenes from the RealEstate10K test set. For fair comparison, all methods use identical input and evaluation viewpoints. The camera poses of the input views are provided, with two views used for reconstruction and three additional views used for evaluation in each scene.

\begin{table}[htbp]
    \caption{Quantitative comparison of reconstruction quality with optimization-based methods. (iter.: number of iterations; con.: connectivity, \textit{i.e.}, whether the generated representation is geometrically connected.)}
    
	\centering
    \normalsize
   \begin{threeparttable}
	\setlength{\tabcolsep}{4pt}
	{\begin{tabular}{*{6}{c}}
	\toprule
        algorithm & PSNR$\uparrow$ & SSIM$\uparrow$ & LPIPS$\downarrow$ & iter. $\downarrow$ & con. \\
	\midrule
        3DGS & 18.35 & 0.601 & 0.350 & 30k & $\times$ \\
        2DGS & 16.17 & 0.526 & 0.419 & 30k & $\times$ \\
        \specialrule{0.3pt}{0pt}{0pt}
        trianglesplatting & 18.53 & 0.683 & \underline{0.279} & 30k & $\times$ \\
        meshsplatting & \underline{19.78} & \underline{0.685} & 0.340 & 30k & $\checkmark$ \\
        FTSplat(Ours) & \textbf{20.39} & \textbf{0.707} & \textbf{0.257} & 1 & $\checkmark$ \\
	\bottomrule
\end{tabular}}
\end{threeparttable}
\label{table_opt}
\end{table}

As shown in Table~\ref{table_opt}, 
our method consistently achieves higher PSNR, SSIM, and LPIPS scores than optimization-based triangle rasterization methods under sparse-view settings, while requiring only a single forward pass. In contrast to optimization-based methods that typically require minutes of per-scene iterative refinement, our approach reconstructs each scene in only 0.17s. Moreover, similar to MeshSplatting, the generated surfaces produced by our method preserve structural connectivity, which further enables the resulting triangular surface models to be directly compatible with existing graphics and robotics simulation platforms without additional post-processing.

As illustrated in Fig.~\ref{fig_triangle}, we further present a qualitative comparison between our method and existing optimization-based triangle rasterization approaches under sparse-view settings. Although optimization-based methods can achieve high reconstruction quality when a large number of training views are available, their performance degrades when only a limited number of views with large viewpoint variations are provided. In such cases, iterative optimization tends to converge to local minima during triangular surface reconstruction, resulting in noticeable noise and structural artifacts when rendering from novel viewpoints. In particular, as the disparity between training views increases, the generalization ability of optimization-based methods to unseen viewpoints becomes increasingly limited. In contrast, our feed-forward framework is able to produce geometrically consistent triangular surface models across viewpoints through a single forward pass, yielding more stable and coherent reconstructions under sparse-view inputs.

\subsection{Reconstruction Quality Comparison with Feed-forward Gaussian Splatting Methods}
In this subsection, we conduct quantitative evaluations to compare our method with existing feed-forward Gaussian Splatting (GS) approaches in terms of novel view synthesis quality.
In addition, we provide qualitative comparisons to assess the geometric consistency of the reconstructed 3D models in the actual 3D space between our method and feed-forward GS methods.
\begin{table}[htbp]
	\caption{Quantitative comparison of reconstruction quality with feed-forward Gaussian Splatting methods.}
	\centering
    \normalsize
   \begin{threeparttable}
	\setlength{\tabcolsep}{4pt}
	{\begin{tabular}{*{4}{c}}
	\toprule
        algorithm &PSNR$\uparrow$&SSIM$\uparrow$&LPIPS$\downarrow$\\
	\midrule
        Mvsplat & \underline{27.03} & \underline{0.891} & \underline{0.106} \\
        Depthsplat & \textbf{27.61} & \textbf{0.903} & \textbf{0.099} \\
        \specialrule{0.3pt}{0pt}{0pt}
        FTSplat(Ours) & 20.39 & 0.707 & 0.257 \\
	\bottomrule
\end{tabular}}
\end{threeparttable}
\label{table_gs}
\end{table}

As shown in Table~\ref{table_gs}, in terms of quantitative novel view rendering performance, the reconstructed triangular surface produced by our method is slightly inferior to the results rendered by feed-forward GS methods. This observation is consistent with the comparison between mesh-based methods and the original 3DGS reported in~\cite{meshsplatting}.
However, regarding 3D spatial consistency, as illustrated in Fig.~\ref{fig_gs}, the triangle surfaces reconstructed by our method effectively eliminates the fog-like floating artifacts commonly observed in 3DGS reconstructions. As a result, our method produces a cleaner and more geometrically consistent 3D representation, which is more suitable for direct use in robotic perception tasks.
\subsection{Ablation Study}
\textbf{Quantitative Comparison.}
We conduct an ablation study to evaluate the effectiveness of the proposed relative 3D point cloud supervision in our feed-forward triangle surfaces generation framework. In particular, we investigate the impact of different point cloud generation methods used for supervision, including VGGT and Depth Anything 3. The quantitative results are summarized in Table~\ref{table_ablation}.

\begin{table}[htbp]
\vspace{2mm}
	\caption{
    Quantitative ablation study on relative 3D point cloud supervision.(w/o: without point cloud supervision; w/: point cloud signals provided by VGGT or Depth Anything 3.)
    }
	\centering
    \normalsize
   \begin{threeparttable}
	\setlength{\tabcolsep}{4pt}
	{\begin{tabular}{*{4}{c}}
	\toprule
        algorithm &PSNR$\uparrow$&SSIM$\uparrow$&LPIPS$\downarrow$\\
	\midrule
        w/o & 13.06 & 0.401 & 0.509 \\
        w/ vggt & \underline{20.07} & \underline{0.692} & \underline{0.275} \\
        w/ da3 & \textbf{20.39} & \textbf{0.707} & \textbf{0.257} \\
	\bottomrule
\end{tabular}}
\end{threeparttable}
\label{table_ablation}
\end{table}

As shown in the table, employing Depth Anything 3 to generate supervisory point clouds leads to the best overall performance. Using VGGT as the external point cloud generator yields slightly inferior results, while removing the relative 3D point cloud supervision results in a significant performance degradation. These findings demonstrate that the proposed supervision strategy plays a critical role in improving reconstruction quality, and that the choice of point cloud generation method further influences the final performance.

\begin{figure}[htbp]
\vspace{2mm}
    \centering
    \begin{minipage}{0.23\textwidth}
        \centering
        \includegraphics[width=\linewidth]{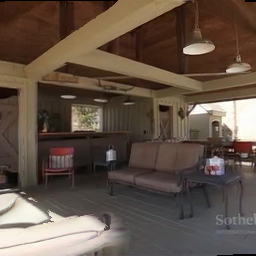}
        
        \vspace{0.5em} 
        {\small (a) w/ Point Supervision} 
    \end{minipage}
    \hfill
    \begin{minipage}{0.23\textwidth}
        \centering
        \includegraphics[width=\linewidth]{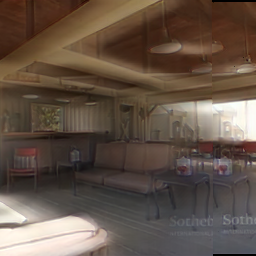}
        
        \vspace{0.5em}
        {\small (b) w/o Point Supervision}
    \end{minipage}
    
    \caption{Qualitative comparison of rendering quality under different ablation settings}
    \label{ab_img}
\end{figure}

\textbf{Qualitative Comparison.}
As illustrated in Fig.~\ref{ab_img}, removing the relative 3D point cloud supervision leads to a noticeable degradation in the rendering quality of the reconstructed triangle surfaces produced by the feed-forward model. In particular, the synthesized novel views exhibit significant blurring and visible artifacts when the supervision is absent. This behavior suggests that, without explicit geometric regularization, the network struggles to learn structurally consistent surface representations during training.

\begin{figure}[htbp]
    \centering
    \begin{minipage}{0.23\textwidth}
        \centering
        \includegraphics[width=\linewidth]{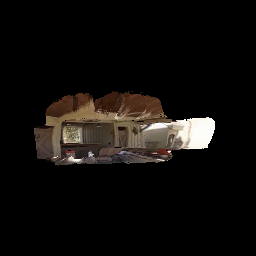}
        
        \vspace{0.5em} 
        {\small (a) w/ Point Supervision} 
    \end{minipage}
    \hfill
    \begin{minipage}{0.23\textwidth}
        \centering
        \includegraphics[width=\linewidth]{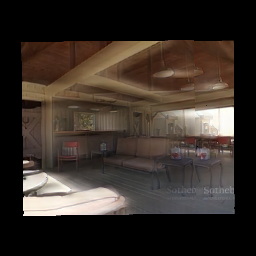}
        
        \vspace{0.5em}
        {\small (b) w/o Point Supervision}
    \end{minipage}
    
    \caption{Qualitative comparison of 3D structure reconstruction under different ablation settings}
    \label{ab_3d}
\end{figure}

More pronounced differences emerge when analyzing the reconstructed 3D geometry itself. As shown in Fig.~\ref{ab_3d}, in the absence of relative 3D point cloud supervision, the model lacks explicit geometric constraints to guide structural recovery. Consequently, although the optimization may appear to converge in the image rendering space, it tends to reach a pseudo-converged solution corresponding to a local minimum. This results in a degenerate 3D reconstruction, where the recovered triangle surface collapses into a nearly planar, image-stitching-like structure rather than forming a geometrically coherent surface.


Both quantitative and qualitative results consistently validate the effectiveness of the proposed relative 3D point cloud supervision. Quantitatively, it improves all evaluation metrics, while qualitatively enhancing rendering fidelity and geometric consistency of the reconstructed surfaces. These findings confirm its critical role in boosting the overall performance of the feed-forward triangle surface generation framework.

\section{CONCLUSIONS}
In this work, we presented FTSplat, a feed-forward framework that directly generates continuous triangular surface representations from multi-view images. By combining efficient single-pass inference with mesh-based geometric structure, the proposed method enables fast and stable 3D reconstruction without per-scene optimization while producing simulation-ready surfaces. The pixel-aligned triangle generation module and relative 3D point-cloud supervision improve geometric consistency, leading to reliable reconstruction under sparse-view settings. Experimental results demonstrate strong reconstruction quality, improved cross-view consistency, and direct compatibility with graphics and robotic simulation platforms.

Despite these advantages, limitations remain in handling occluded regions, where incomplete geometric cues may degrade surface estimation. Future work will explore more robust surface generation strategies and stronger geometric priors to further enhance reconstruction performance in challenging scenarios.



\addtolength{\textheight}{-12cm}   









\end{document}